\documentclass{article}
\usepackage{ijcai24}

\usepackage{times}
\usepackage{soul}
\usepackage{url}
\usepackage[hidelinks]{hyperref}
\usepackage[utf8]{inputenc}
\usepackage[small]{caption}
\usepackage{graphicx}
\usepackage{amsmath}
\usepackage{amsthm}
\usepackage{booktabs}
\usepackage{algorithm}
\usepackage[switch]{lineno}
\usepackage{float}
\usepackage{amssymb}
\usepackage{multirow}

\usepackage[capitalize]{cleveref}
\crefname{section}{Sec.}{Secs.}
\Crefname{section}{Section}{Sections}
\Crefname{table}{Table}{Tables}
\crefname{table}{Tab.}{Tabs.}

\usepackage{algorithmicx}
\usepackage{algpseudocode}

\usepackage{mathtools}

\title{Why Only Text: Empowering Vision-and-Language Navigation with\\ Multi-modal Prompts}

\author{
Haodong Hong$^{1,2}$
\and
Sen Wang$^{1*}$\and
Zi Huang$^{1}$\and
Qi Wu$^3$\And
Jiajun Liu$^{2,1}$\thanks{Corresponding author}\\
\affiliations
$^1$The University of Queensland\\
$^2$CSIRO Data61\\
$^3$The University of Adelaide\\
\emails
\{haodong.hong, sen.wang\}@uq.edu.au,
huang@itee.uq.edu.au,
qi.wu01@adelaide.edu.au,
jiajun.liu@csiro.au
}

\begin{document}

\maketitle

\begin{abstract}
Current Vision-and-Language Navigation (VLN) tasks mainly employ textual instructions to guide agents. 
However, being inherently abstract, the same textual instruction can be associated with different visual signals, causing severe ambiguity and limiting the transfer of prior knowledge in the vision domain from the user to the agent. 
To fill this gap, we propose Vision-and-Language Navigation with Multi-modal Prompts (VLN-MP), a novel task augmenting traditional VLN by integrating both natural language and images in instructions.
VLN-MP not only maintains backward compatibility by effectively handling text-only prompts but also consistently shows advantages with different quantities and relevance of visual prompts. 
Possible forms of visual prompts include both exact and similar object images, providing adaptability and versatility in diverse navigation scenarios.
To evaluate VLN-MP under a unified framework, we implement a new benchmark that offers:
(1) a training-free pipeline to transform textual instructions into multi-modal forms with landmark images;
(2) diverse datasets with multi-modal instructions for different downstream tasks; 
(3) a novel module designed to process various image prompts for seamless integration with state-of-the-art VLN models.
Extensive experiments on four VLN benchmarks (R2R, RxR, REVERIE, CVDN) show that incorporating visual prompts significantly boosts navigation performance.
While maintaining efficiency with text-only prompts, VLN-MP enables agents to navigate in the pre-explore setting and outperform text-based models, showing its broader applicability.
Code is available at \url{https://github.com/honghd16/VLN-MP}.
\end{abstract}

\section{Introduction}

\begin{figure}[ht]
  \centering
  \includegraphics[width=\linewidth]{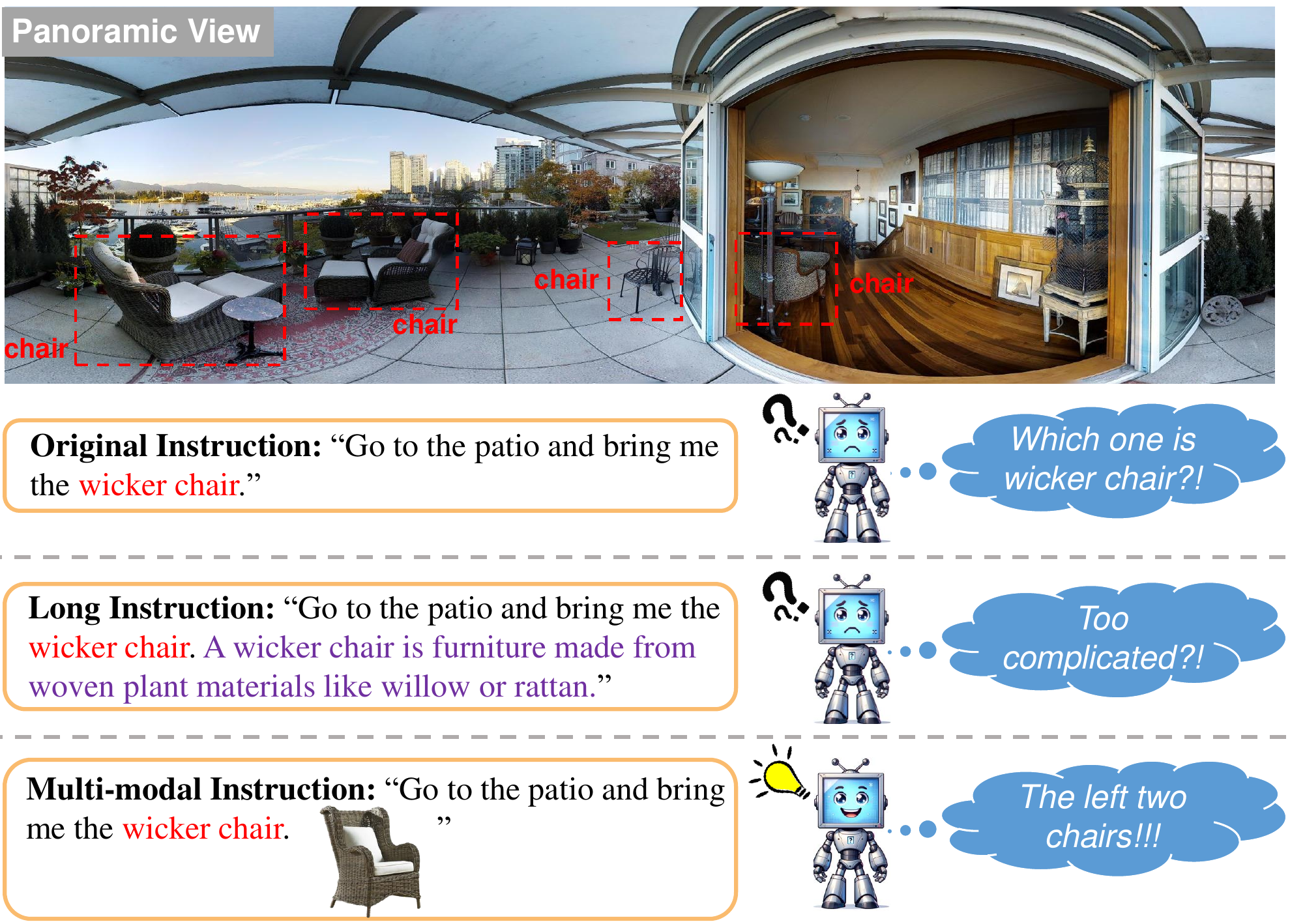}
  \caption{An example of the effectiveness of multi-modal prompts for VLN. Initially, the agent struggles to identify the ``wicker chair'' due to unfamiliarity. While detailed textual explanations can increase comprehension challenges, introducing a visual prompt from the web is convenient and simplifies the task significantly. In this case, a photo of a wicker chair provides a clear, direct reference to help the agent accurately identify the target object.}
  \label{fig:example}
\end{figure}

With the booming of large language models (LLMs)~\cite{brown2020language,touvron2023llama}, there is a growing interest in employing natural language to guide embodied agents in executing various tasks~\cite{huang2022language}.
Vision-and-Language Navigation (VLN)~\cite{anderson2018vision} is a representative example, demanding not only individual language and vision understanding but also the co-grounding between these two modalities.
Numerous methods~\cite{zhu2020vision,zhu2023vision,wang2023dual,majumdar2020improving,wang2023scaling,zhou2024navgpt} and datasets~\cite{jain2019stay,krantz2020beyond,qi2020reverie,ku2020room} have been proposed to address different challenges in VLN.

Current VLN tasks mainly rely on natural language instructions.
However, real-world navigation often involves scenarios where images are crucial as guidance, providing essential and convenient supplementary information to instructions, as shown in \cref{fig:example}.
These image-enhanced instructions reduce the reliance on perfect language instructions and provide more precise path or destination depiction, thus helping agent navigation.
For example, a photo of the target object can effectively supplement the language descriptions that might be ambiguous. 
Moreover, while detailed language instructions are challenging to formulate, combining simple environmental photos with brief descriptions can create efficient multi-modal guidance.
Such multi-modal prompts are standard practices in human navigation, whether when using apps like Google Maps or capturing images for complex descriptions.
The ubiquity of cameras and smartphones has made acquiring these images practical and accessible, yet this setting remains under-explored in existing VLN research.

To address this, we introduce a novel task, Vision-and-Language Navigation with Multi-modal Prompts (VLN-MP), in which agents navigate with instructions that combine visual signals and natural language.
Considering the varying quantity and quality of visual prompts in applications, we propose three distinct settings—Aligned, Related, and Terminal—to accommodate different scenarios. 
The Aligned setting needs multiple actual images, while the other two settings only demand similar views or a single view of the destination.
Agents in VLN-MP are expected to be versatile across diverse image prompt modes, covering these three settings and handling varying numbers of images.
VLN-MP works as an advancement over traditional VLN, enabling models to still operate effectively with textual instructions alone, yet achieving enhanced performance when supplemented with various image prompts, even a single image.

To facilitate research in VLN-MP, we propose a new benchmark with several essential components.
Recognizing the importance and prevalence of landmarks in navigation~\cite{yesiltepe2021landmarks,he2021landmark}, we focus on landmarks within instructions as the primary source of visual prompts. 
To transform textual instructions into multi-modal forms, we introduce a universal data generation pipeline leveraging large pre-training models~\cite{openai2023gpt4,li2022grounded} to realize landmark phrase extraction, landmark detection, image selection, and data augmentation. 
Applying this pipeline to existing VLN datasets, we generate their corresponding multi-modal instruction datasets with various settings to assess agent adaptability to different image prompt modes.
Furthermore, we develop a novel module, Multi-modal Prompts Fusion (MPF), which integrates features from visual prompts and text tokens in a unified embedding space.
MPF is compatible with state-of-the-art VLN models like HAMT~\cite{chen2021history} and DUET~\cite{chen2022think} for enhanced navigation performance.

We conduct extensive experiments on four datasets—R2R, RxR, REVERIE, and CVDN—to demonstrate that agents trained in VLN-MP improve navigation performance across different image prompt settings while maintaining robustness in traditional VLN tasks. 
Through these experiments, we also provide valuable insights on leveraging multi-modal instructions to boost agent generalizability.
Additionally, training agents under the VLN-MP paradigm prepares them for application in the pre-explore setting of VLN, where they outperform agents trained under the original setup.

\begin{figure*}[t]
  \centering
  \includegraphics[width=\textwidth]{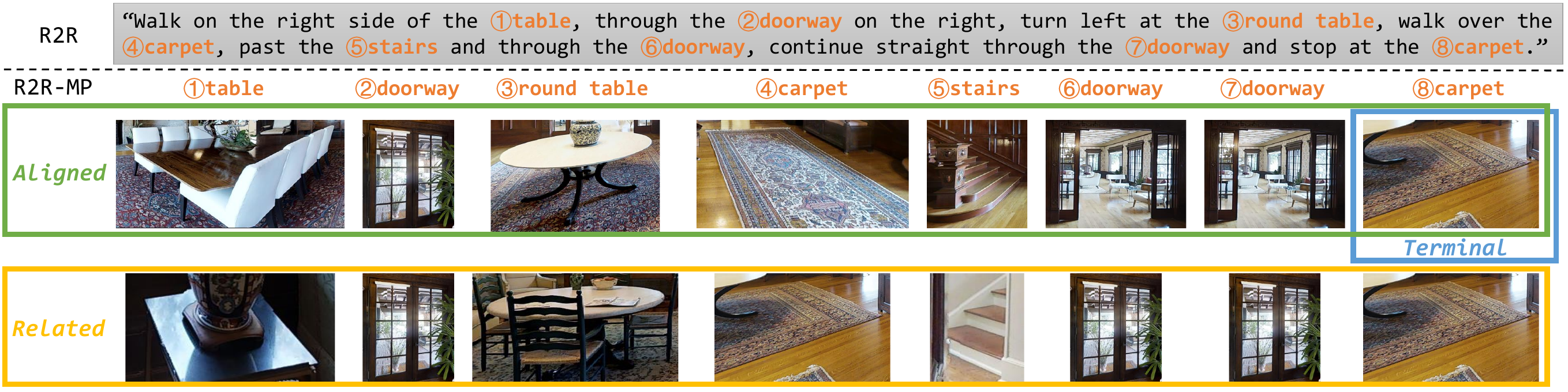}
  \caption{An example of the R2R instruction and the landmark images in three settings of R2R-MP. The aligned setting (green) provides sufficient highly grounded landmark images. The related setting (yellow) presents similar images associated with the phrase but not necessarily the instruction, like the same image for two carpets. The Terminal setting (blue)  includes only an image of the last landmark. }
  \label{fig:dataset}
\end{figure*}

\section{Related Work}
\paragraph{Vision-and-Language Navigation.} 
In VLN, agents navigate to a target destination following natural language instructions within simulated environments such as Matterport3D~\cite{chang2017matterport3d}.
Since the introduction of the Room-to-Room (R2R) dataset~\cite{anderson2018vision}, numerous tasks and datasets have been developed to address various challenges and scenarios for instruction following agents.
Ku \textit{et al.}~\cite{ku2020room} proposed Room-across-Room (RxR), a multilingual dataset featuring fine-grained instructions and pose traces of annotators.
REVERIE~\cite{qi2020reverie} was presented in which instructions only describe the target location and object rather than providing step-by-step guidance.
Thomason \textit{et al.}~\cite{thomason2020vision} developed the first vision-and-dialogue navigation dataset, CVDN, which requires agents to learn from the dialogue history and ask for assistance during navigation.
Liu \textit{et al.}~\cite{liu2023aerialvln} extends the VLN setting from ground scenarios to the sky and proposes the AerialVLN task with UAV-based environments.
However, these works are confined to text-only instructions, while our work serves as the first paradigm to facilitate the use of multi-modal instructions in VLN.

\paragraph{Landmarks in VLN.}
Previous research has highlighted the significance of landmarks and extensively employed them in navigation agents~\cite{berg2020grounding,yesiltepe2021landmarks}.
In VLN, landmarks have been used for navigation graph construction~\cite{hong2020language}, instruction and trajectory decomposition~\cite{he2021landmark}, and instruction generation~\cite{wang2022less}.
GELA~\cite{cui2023grounded} introduces the human-annotated entity-landmark dataset GEL-R2R to enhance cross-modal alignment of agents.
Wang \textit{et al.}~\cite{wang2022less} also incorporated landmark images to generate multi-modal instructions.
However, our work differs in two key ways.
Firstly, while they aim at producing novel textual instructions, we emphasize multi-modal instructions as the direct prompt for navigation.
Secondly, unlike their strict requirements, our work allows for a broader range of landmark image quality and quantity.

\paragraph{Multi-modal Prompts.}
Multi-modal prompts have gained increasing popularity for their superior expressive power over single-modality and uniform input interfaces for training Visual Language Models (VLM).
Flamingo~\cite{alayrac2022flamingo} leverages large-scale multi-modal web corpora containing interleaved text and images to complete various textual tasks with few examples.
VIMA~\cite{jiang2022vima} establishes the first robot learning benchmark for multi-modal-prompted tasks, providing corresponding models, datasets, and evaluation protocols.
MaPLe~\cite{khattak2023maple} incorporates multi-modal prompt learning into CLIP~\cite{radford2021learning} to enhance model generalization, though its focus remains on classification tasks and can not manage complex instructions or detailed panoramas.
Our work contributes to the VLN task, where multi-modal instructions are rarely studied but hold significant importance.

\section{VLN-MP Task}
\subsection{Task Definition}
In VLN, an agent needs to navigate towards a target destination following a natural language instruction $\hat{x}=(x_1,x_2,...,x_L)$ with $L$ words.
At each timestep $t$, the agent observes its surroundings through a panoramic representation comprising $N=36$ views $O_t=\left\{o_t^i\right\}_{i=1}^{N}$. 
Each view $o_t^i$ includes an RGB image $I_t^i$ and directional details of heading $\theta_t^i$ and elevation $\phi_t^i$ angles of its current node $V_t$.
Based on its policy $\pi$, the agent decides on an action $a_t$ to transition to one of the neighboring nodes $\mathcal{N}(V_t)$ by selecting the view that aligns best with the target node.
In this process, visual and textual signals are separate: the visual ones constantly change during navigation, while the textual input remains static, serving as the human-agent interface. 
However, in real-world navigation, images are often employed as part of the instructions, providing essential supplementary information.
This problem is ignored by current VLN works, complicating human-agent communication and preventing agents from using image prompts to improve performance.

Therefore, we propose the Vision-and-Language Navigation with Multi-modal Prompts (VLN-MP) task first to consider multi-modal instructions in navigation.
Assume we provide $n$ images $(I_1, I_2, ..., I_n)$ corresponding to $n$ phrases $(z_1,z_2,...,z_n)$ within instruction $\hat{x}$. 
Each phrase $z_i$ is represented as $z_i=x_{start_i:end_i}$, where $x_{i:j} = (x_i, x_{i+1},..., x_j)$ denotes words from position $i$ to $j$.
By inserting images next to corresponding phrases, the textual instruction $\hat{x}$ is transformed into the multi-modal form:
\begin{equation}
    \Bar{x} = (x_{1:end_1}, I_1, x_{end_1+1:end_2}, I_2,..., I_n, x_{end_n+1:L})
\end{equation}
The agent receives $\Bar{x}$ instead of $\hat{x}$ as navigation guidance, with others the same as in VLN. 
Here we focus on scenarios where each phrase is paired with a single image.

\subsection{Image Prompt Settings}
\label{sec:ips}

Due to varying difficulties in acquiring images, agents may be provided with visual prompts of different numbers and qualities, ranging from one similar image from the web to sufficient actual images from cameras.
To enable agents to adapt to such variations, we propose three different settings for VLN-MP based on the number and relevance of the prompt images in the instructions: \textbf{Aligned}, \textbf{Related}, and \textbf{Terminal}.
The Aligned setting aims to provide precise and abundant prompts with two criteria. 
Firstly, each image $I_i$ should be aligned with the actual view that the phrase $z_i$ describes.
Secondly, the number of images $n$ should be large enough to cover the entire instruction. 
The Related setting relaxes the first requirement and only demands $I_i$ related to the phrase, aiming to facilitate instruction understanding without requiring perfect alignment.
The Terminal setting removes the second requirement, providing only one image depicting the view around the destination.  
This setting helps agents target the stop position and accomplish object-centric tasks such as object navigation~\cite{zhu2021soon}.
Besides these settings, we also randomly vary the number of visual prompts to further evaluate agent adaptability.

\begin{figure}[t]
  \centering
  \includegraphics[width=\linewidth]{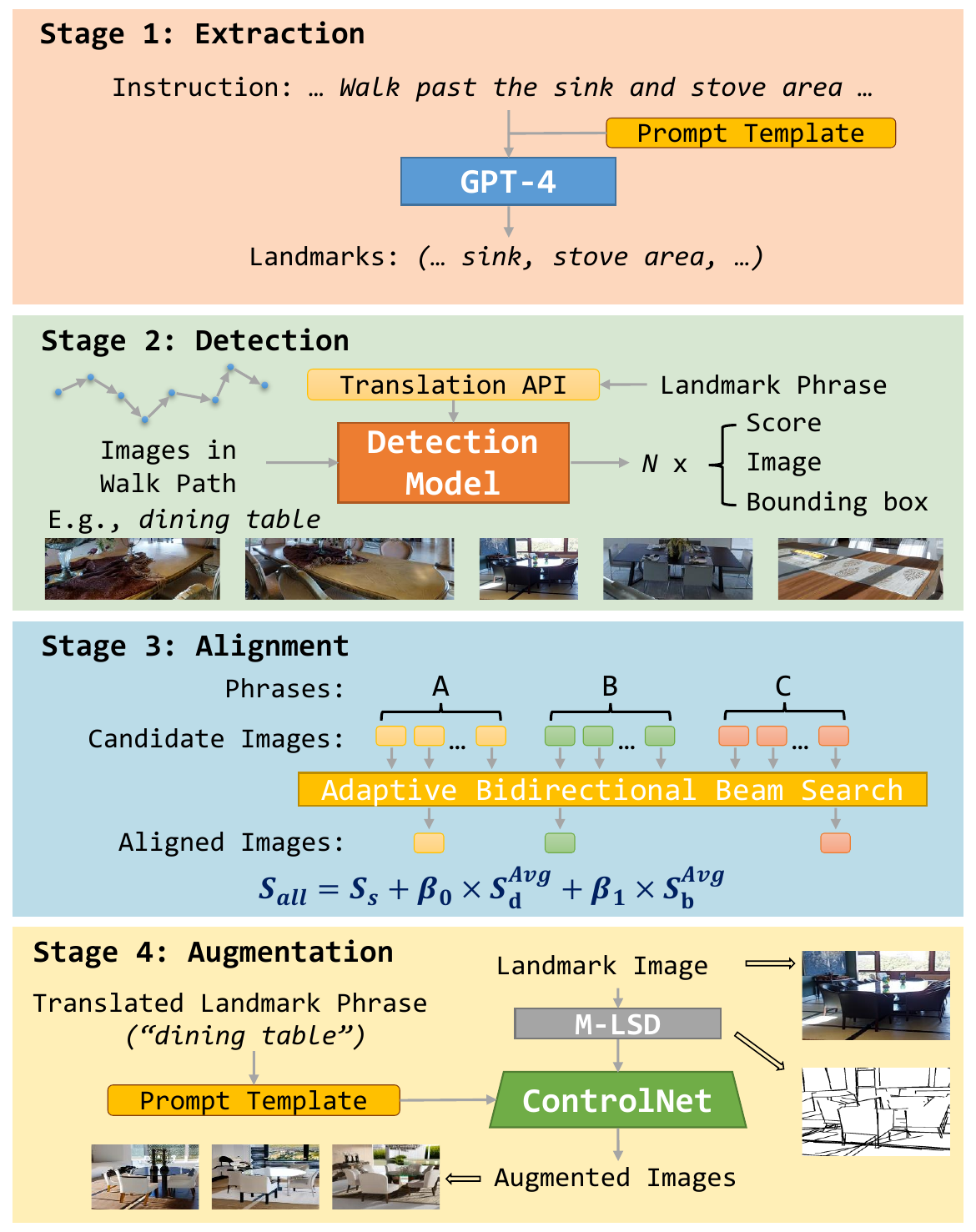}
  \caption{Four stages of our data generation pipeline for transforming textual instructions into multi-modal forms.}
  \label{fig:pipeline}
\end{figure}

\section{Benchmark Implementation}

In this section, we introduce the proposed benchmark of VLN-MP in detail, including the data generation pipeline, four novel datasets, and the MPF module.

\subsection{VLN-MP Data Generation Pipeline}

To utilize textual instructions in VLN, we propose a novel data generation pipeline to transform them into multi-modal forms, as shown in \cref{fig:pipeline}. 
Using large pre-training models, our pipeline can be applied to VLN datasets both generally and efficiently without extra training.

\paragraph{Extraction.}
Extracting landmark phrases from instructions is a crucial first step, as its results will impact all subsequent processes.
Fortunately, large language models excel in this task.
Previous work~\cite{wang2022less} employed a BERT-based parser~\cite{kenton2019bert} but still misidentified irrelevant words as landmarks.
Moreover, some landmark phrases are not suitable for detection within environments. 
To improve accuracy and accommodate subsequent stages, we utilize the advanced GPT-4 model~\cite{openai2023gpt4} from OpenAI with specially crafted prompts to perform the extraction.  

\paragraph{Detection.}
With the landmark phrases, we detect potential entities in the environment, a task requiring phrase grounding or open-vocabulary object detection.
Our strategy involves selecting different models based on the instruction type. 
For object-centered instructions or short instructions, we use GroundingDINO~\cite{liu2023grounding}, known for its high accuracy in detecting specific object names.
For detailed and lengthy instructions, we choose the GLIP~\cite{li2022grounded} model, which excels in phrase grounding and is more suitable for objects contextualized within phrases.
Specifically, we use the chosen model in a zero-shot manner to scan all images along the path $(N_1, N_2, ..., N_m)$ consisting of $m$ nodes, and generate multiple candidates for each landmark phrase $z_i$.
The resulting candidates $C$ are represented in triads $C = (S, I, B)$, where $S$ is the detection score, $I$ is the detected image, and $B$ indicates the bounding box.

\paragraph{Alignment.}
In this stage, we select the most suitable image from numerous candidates for each phrase based on different settings. 
In the Related setting, we choose the image with the highest detection score by ${i}^{*}=\arg \max _{j \in\left[1, k_{i}\right]}S_{j}^{i}$ for $z_i$, corresponding to the scenario of choosing a similar picture from the surroundings or the web.
For the Aligned setting, we use the order of landmarks in the instruction to guide our selection, prioritizing images that appear in the same sequence. 
We apply the Kendall rank correlation coefficient to measure this alignment as the sequence score $S_s = \frac{4 \times c}{n \times (n - 1)} - 1$, where $c$ is the number of concordant pairs ($I_i \leq I_j$ for any $1 \leq i < j \leq n$).
Additionally, we introduce the bounding box score $S_b$ to prioritize larger images for more details.
The final score for selecting images is a combination of detection score $S_d$, sequence score $S_s$, and bounding box score $S_b$:
\begin{equation}
   S_{all} = S_s + \beta_0 \times S_d^{Avg} + \beta_1 \times S_b^{Avg}
   \label{eq: score}
\end{equation}
For lengthy instructions, we propose an adaptive bidirectional beam search to find the best image combinations from vast search space efficiently.
For the Terminal setting, we directly use the image of the last landmark from the Aligned setting. 

\paragraph{Augmentation.}
A notable advantage of visual prompts is the ease and diversity of data augmentation compared to natural language.
Unlike previous studies~\cite{liu2021vision,li2022envedit}, we employ ControlNet~\cite{zhang2023adding} to generate $N_{aug}$ novel images based on the line segments of the landmark images from M-LSD~\cite{gu2022towards}.
This operation aims to maintain the coarse-grained features like shape and layout consistent with the visual observations while altering fine-grained details like texture and color to enrich visual prompts.
These images serve as augmented multi-modal instructions during training to enhance the dataset diversity and improve the model's generalizability.

\subsection{Datasets for VLN-MP}

We apply our pipeline to four popular VLN datasets—R2R~\cite{anderson2018vision}, RxR~\cite{ku2020room}, REVERIE~\cite{qi2020reverie}, and CVDN~\cite{thomason2020vision}—generating their corresponding multi-modal versions, denoted with the suffix ``\textit{-MP}''.

\paragraph{R2R-MP.} R2R includes 17,409 instructions~\footnote{We did not include the test split with paths not open source.} for short paths in the Matterport3D dataset~\cite{chang2017matterport3d}. 
Each instruction details the route from start to finish and contains several landmark phrases.
Using GroundingDINO, our pipeline produces 17,328 multi-modal instructions, with only 0.4\% of instructions failing to detect any landmark images.
The average number of landmarks for the Aligned setting is 4.15, reaching up to 18 in some instructions.
An example of the three settings in R2R-MP is shown in \cref{fig:dataset}.

\paragraph{RxR-MP.} RxR provides 101,932 instructions with three languages, longer paths, more visible entities, and time alignment between words and virtual poses.
We use our pipeline with the GLIP model to generate 100,923 multi-modal instructions per setting.
The Aligned setting has an average of 7.17 landmarks, with a maximum of 63. 
Additionally, we integrated the Marky-mT5 dataset~\cite{wang2022less} as an alternative Aligned setting for comparison, which has an average of 9.57 landmarks with a maximum of 83 due to the forceful allocation of a frame for each phrase. 

\paragraph{REVERIE-MP.}
Unlike R2R and RxR, REVERIE focuses on object-oriented instructions that describe the target object, typically involving a single landmark—the target object.
We directly use its bounding box annotations to generate 15,410 multi-modal instructions, the same number as REVERIE, with each instruction including one visual prompt.

\paragraph{CVDN-MP.}
CVDN also provides object-oriented instructions but differs from REVERIE by employing ambiguous objects that can refer to multiple instances to make agents use the dialogue history.
In CVDN-MP, we discard the dialogue to emphasize the necessity of visual prompts in ambiguous instructions like ``Find the $\langle$ object $\rangle$'', which is more applicable to real-world situations. 
We also employ BLIP2~\cite{li2023blip} to generate captions for the provided visual prompts to offer an alternative, detailed instruction for comparison. 
CVDN-MP includes 6,031 instructions, each with a visual prompt and an alternate extended instruction.


\subsection{Multi-modal Prompts Fusion Module}

\begin{figure}[t]
  \centering
  \includegraphics[width=\linewidth]{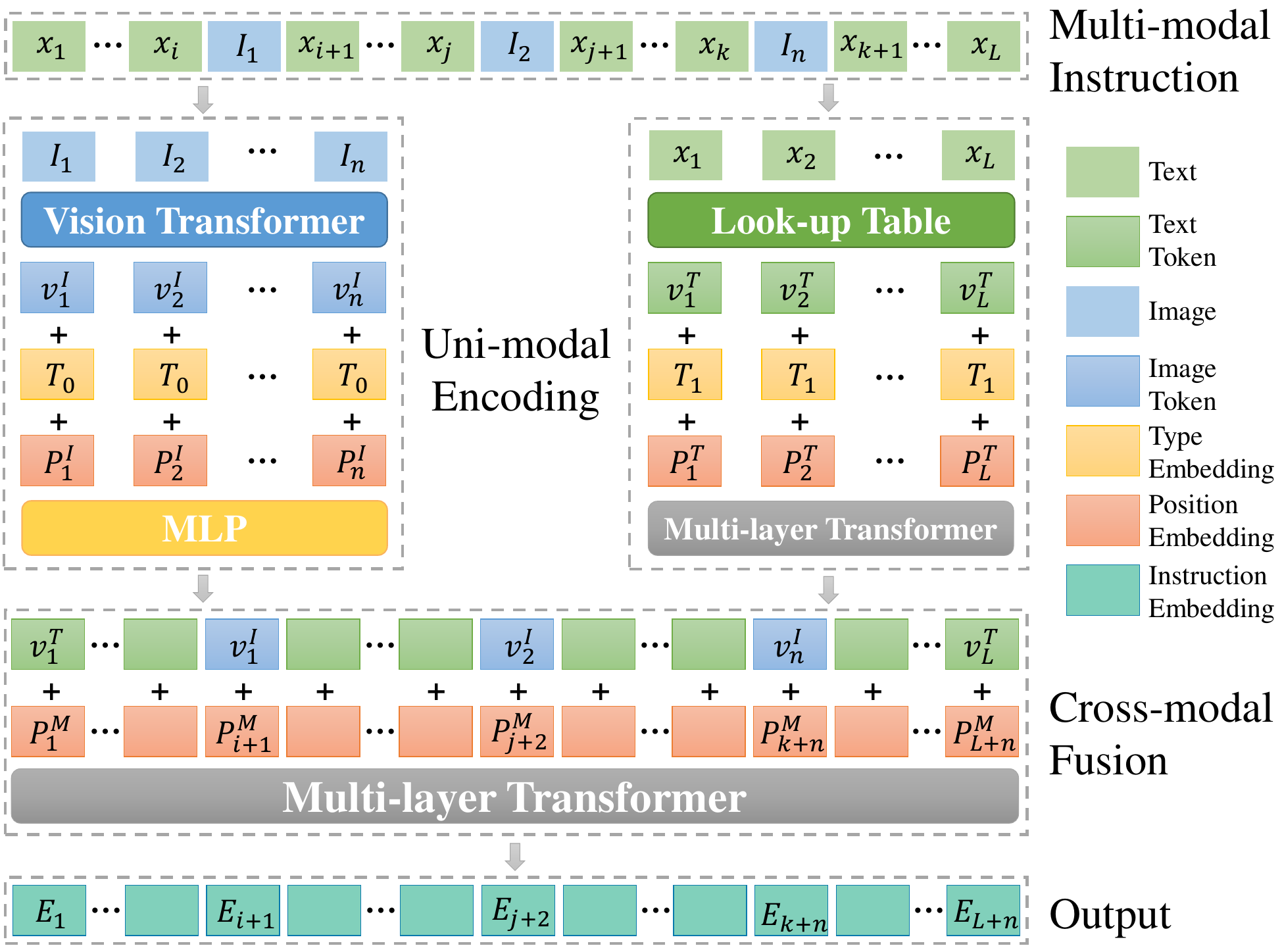}
  \caption{The architecture of the Multi-modal Prompts Fusion (MPF) module. Texts and images are processed separately and then combined to obtain cross-modal representations. Some symbols are omitted for simplicity. }
  \label{fig:network}
\end{figure}

To effectively utilize multi-modal instructions, we propose a novel Multi-modal Prompts Fusion (MPF) module, as illustrated in \cref{fig:network}.
The MPF module includes a visual branch to process visual prompts individually, parallel to the text branch that handles language prompts, and a fusion layer to combine these two types of tokens.
The visual prompts are initially processed by an image encoder to extract features and then enriched with type embedding and position embedding to yield image tokens. 
Concurrently, text tokens are processed and merged with visual tokens, further integrating another layer of position embedding to link them effectively. 
This combined multi-modal token sequence is input into a multi-layer Transformer~\cite{vaswani2017attention} to synthesize the final tokens for the multi-modal instruction. 
These tokens are then further processed to determine the next action according to the design of different VLN models.
Crucially, position encoding is applied twice for each token: the first to mark the sequence of the images and texts separately and the second to match these two types of tokens.
This dual application ensures the model recognizes the relationship between phrases and their corresponding images.

\section{Experiments}

\subsection{Experimental Setup}
\paragraph{Datasets.}
We conduct our experiments on the proposed four VLN-MP datasets.
The original datasets have four splits: train, validation seen (val seen), validation unseen (val unseen), and test unseen.
As the ground truth paths in the test split are not released, their multi-modal versions only contain the first three splits for training and evaluation.
The distinction between val seen and val unseen is that houses in the former are also in the train split~\cite{zhang2021diagnosing}.

\paragraph{Baseline VLN Models.}
We adopt HAMT~\cite{chen2021history} and DUET~\cite{chen2022think} as baseline models, which are the mainstream architectures in VLN. 
HAMT employs a transformer-based network to encode instructions, visual observations, and history together for action prediction. 
DUET expands on HAMT by building a real-time topological map to enable global action decisions over local movement.



\paragraph{Evaluation Metrics.}
We evaluate the agent performance on the following metrics: (1) Success Rate (\textbf{SR}): the ratio of agents stopping within 3 meters of the target; (2) Success rate weighted by Path Length (\textbf{SPL})~\cite{anderson2018evaluation}: SR normalized by the ratio between the length of the shortest path and the predicted path; (3) the normalized Dynamic Time Warping (\textbf{nDTW})~\cite{ilharco2019general}: a measure of instruction fidelity by computing the similarity between the reference path and the predicted path; (4) Goal Progress (\textbf{GP})~\cite{thomason2020vision}: metric for CVDN which measures the average difference between the length of the completed trajectory and the remaining distance to the goal.

\begin{table}[t]
    \centering
    \begin{tabular}{l|ccc}
        \hline
        Score    & Precision & Recall & F1 \\
        \hline
        Fuzzy Matching       & 0.95      & 0.89   & 0.91 \\
        ROGUE-L              & 0.87      & 0.80   & 0.81 \\
        \hline
    \end{tabular}
    \caption{Phrase similarity between R2R-MP and GELR2R.}
    \label{tab:phrase}
\end{table}

\begin{table}[t]
    \centering
    \begin{tabular}{l|cc}
        \hline
        Method    & Matching & Neighboring \\
        \hline
        w/o Alignment       & 0.40      & 0.62    \\
        w/ Alignment        & 0.69      & 0.91   \\
        \hline
    \end{tabular}
    \caption{Accuracy of viewpoint matching and neighboring for R2R-MP and GELR2R images, with and without Alignment stage.}
    \label{tab:align}
\end{table}

\paragraph{Implementation Details.}
In the pipeline, we utilize GPT-4 from OpenAI's official API, and the GLIP-L and GroundingDINO-T models for landmark detection.
The weights for score balancing in \cref{eq: score} are $\beta_0 = 0.5$ and $\beta_1 = 0.1$ to prioritize the sequence score $(S_s)$ over the others, reflecting their relative importance in our method. 
The small $\beta_1$ is used to ensure that the bounding box scores become significant only when multiple bounding boxes are generated for the same object, in which case $S_d$ and $S_s$ remain consistent. 
For non-English languages, we use the Google translate service\footnote{https://cloud.google.com/translate} to translate them into English.
We generate five novel images per visual prompt using the control\_sd15\_mlsd model for data augmentation.
For baseline models, we follow the implementation details in their official repositories.
Following their setting for processing visual observations, we employ the ViT-B/32 model as the feature extractor for landmark images in R2R, CVDN, and REVERIE, while CLIP-ViT-B/16 is used for RxR.
During training, we select the augmented data with a probability of $\gamma=0.2$ to replace original landmark images.
All models are fine-tuned for 200K iterations with a learning rate of 1e-5 and a batch size of 8 on a single NVIDIA A6000 GPU.
The best model is selected based on performance in the val unseen split.

\begin{table*}
\centering
\begin{tabular}{c|p{140pt}|cc|c|ccc|ccc}
\toprule
\multirow{2}{*}[-0.5ex]{\#} & \multirow{2}{*}[-0.5ex]{Model} & \multicolumn{2}{c|}{Image Prompt}  & \multirow{2}{*}[-0.5ex]{Setting} & \multicolumn{3}{c|}{\textbf{Validation Seen}} & \multicolumn{3}{c}{\textbf{Validation Unseen}}\\
\cmidrule(lr){3-4}
\cmidrule(lr){6-8}
\cmidrule(lr){9-11}
 &    & Train & Eval &  & SR$\uparrow$  & SPL$\uparrow$   & nDTW$\uparrow$  & SR$\uparrow$   & SPL$\uparrow$   & nDTW$\uparrow$ \\
\hline
1 & Multi Baseline~\cite{ku2020room} & $\times$  & $\times$ & -     & 25.2  & -      & 42.2     & 22.8   &  -     & 38.9       \\
2 & Mono Baseline~\cite{ku2020room}  & $\times$  & $\times$       & -      & 28.8  & -      & 46.8     & 28.5   &  -     & 44.5       \\
3 & EnvDrop~\cite{shen2021much}      & $\times$  & $\times$       & -     & -      & -           & 42.6   &  -     & 55.7          \\
4 & CLEAR~\cite{li2022clear}         & $\times$  & $\times$       & -     & -      & -         & 44.4   &  39.3  & 57.0         \\
5 & HOP+~\cite{qiao2023hop+}         & $\times$  & $\times$       & -     & 53.6     & 47.9      & 59.0        & 45.7   &  38.4  & 52.0         \\
6 & VLN-PETL~\cite{qiao2023vln}      & $\times$  & $\times$       & -     & 60.5     & 56.8      & 65.7        & 57.8   &  54.2  & 64.9         \\

\hline
7 & HAMT~\cite{chen2021history}      & $\times$ & $\times$  & -   & 59.4   & 58.9  & 65.3     & 56.5   & 56.0   &  63.1   \\ 
8 & HAMT+MPF  & $\checkmark$   & $\times$  & -  & 64.6   & 60.8   &  68.6     & 57.6   & 53.5   &  64.0         \\
9 & HAMT+MPF  & $\checkmark$ & $\checkmark$ & T & 65.8   & 62.0   &  69.2     & 58.1   & 54.0   &  64.6         \\
10 & HAMT+MPF  & $\checkmark$ & $\checkmark$ & R   & 67.2   & 63.0   & 70.1     & 58.7   & 54.1    &  63.6          \\
11 & HAMT+MPF  & $\checkmark$ & $\checkmark$ & A-P  & 68.8   & 63.7     &  71.1     & 59.3   & 55.1     & 65.0    \\
12 & HAMT+MPF  & $\checkmark$ & $\checkmark$ & A-M & \textbf{69.2}  & \textbf{65.5}  & \textbf{71.7}   & \textbf{60.0}  & \textbf{56.3}  & \textbf{66.4}     \\
\bottomrule
\end{tabular}
\caption{Navigation performance on the RxR-MP dataset. ``A'' means Aligned, ``R'' is Related, and ``T'' refers to Terminal; ``P'' and ``M'' indicate images from our pipeline and Marky-mT5, respectively.}
\label{tab:RxR-MP}
\end{table*}

\subsection{Dataset Evaluation}
We compare the extracted phrases and landmark images with Marky-mT5 and GELR2R datasets for evaluation.

\paragraph{Phrases.} Given the multiple errors produced by the BERT-based parser in Marky-mT5, we mainly compare our phrases with the human-annotated data from GELR2R. 
To thoroughly assess the similarity of phrases between our R2R-MP and GELR2R, we adopt Fuzzy Matching and ROGUE-L~\cite{lin2004rouge} scoring and calculate the Precision, Recall, and F1 scores for each method.
As shown in ~\cref{tab:phrase}, our results achieve high similarity with the gold data in GELR2R, especially with an impressive 95\% precision in fuzzy matching.
This result demonstrates the effectiveness of our method.

\begin{figure}[ht]
  \centering
  \includegraphics[width=\linewidth]{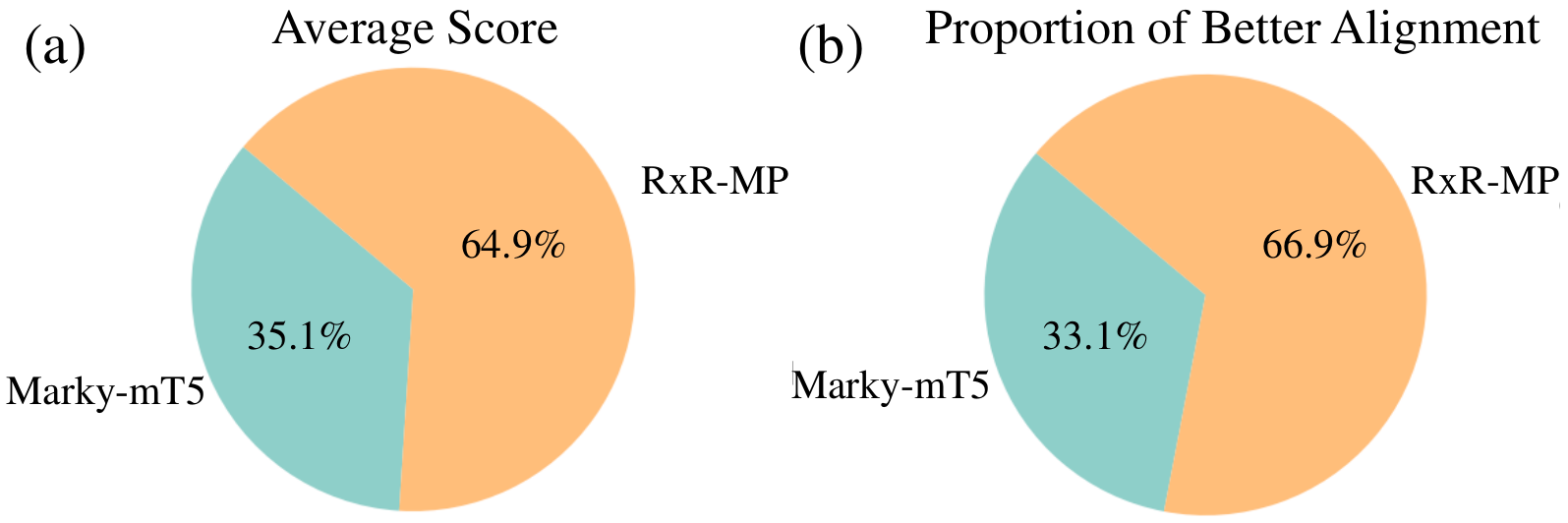}
  \caption{The average score and proportion of a better alignment of Marky-mT5 and our RxR-MP towards the landmark phrases.}
\label{fig:clip}
\end{figure}

\paragraph{Images.} For landmark images, we first prove that our object-centric approach offers enhanced accuracy over Marky-mT5 in the RxR dataset. 
We employ the CLIP model~\cite{radford2021learning} to measure the alignment of images from both datasets with the landmark phrases in a binary classification manner. 
The results, as depicted in \cref{fig:clip}, reveal that RxR-MP outperforms Marky-mT5 with an average score of 0.649 compared to 0.351. 
Furthermore, we also quantify the number of instances where the phrase demonstrates a preference for either dataset. 
Additionally, 66.9\% of phrases favored images from RxR-MP over 33.1\% for Marky-mT5, indicating a stronger alignment of our images with the landmark phrases. 

To demonstrate the effectiveness of our alignment phase, we calculate the matching accuracy of the viewpoints of landmark images in R2R-MP with the ground truth in GELR2R, including the exact matching and neighboring viewpoints. 
The comparison also included images with the highest detection scores without alignment in ~\cref{tab:align}. 
Nearly 70\% of landmark phrases in R2R-MP correctly identify the corresponding viewpoints, and 91\% are adjacent to the actual viewpoint, which may also provide a different view for the same landmark, far surpassing those without alignment, underscoring the significance of our method. 

\subsection{Navigation Performance}
We provide the navigation performance of RxR-MP and CVDN-MP here to represent the step-wise instructions and goal-oriented instructions.

\paragraph{RxR-MP.}
\cref{tab:RxR-MP} presents the performance of different models on the val seen and unseen splits of the RxR-MP dataset.
Agents in experiments \#1-7 are trained on text-only instructions, and our HAMT+MPF agent in experiments \#8-12 is trained with multi-modal instructions and evaluated under different settings.
Experiments \#7 and \#8 demonstrate that VLN-MP-trained models maintain backward compatibility and achieve higher performance with textual instructions, especially in seen scenarios. 
Comparing experiments \#9-12, we can see that our agent adapts well to different visual prompts and consistently outperforms text-only scenarios, even with one single image.
It also reveals that the agent performance improves with the increase in the number and relevance of visual prompts, which is intuitive and outlines our alignment stage.
The Marky-mT5 version achieves the best results, as it provides larger and more images.

\paragraph{CVDN-MP.} In CVDN-MP, we investigate the situation when the visual prompts provide necessary information for navigation.
\cref{tab:cvdn} shows the Goal Progress (GP) of agents when presented with various types of prompts.
Compared with single-modality prompts, utilizing multi-modal prompts maintains backward compatibility and significantly performs better with a target image.
This phenomenon highlights the significance of visual prompts in goal-oriented tasks and the effectiveness of our MPF module.
The captions of the visual prompts enhance GP in seen scenarios but not in unseen ones, suggesting that visual prompts are more intuitive and effective than language.

\begin{table}
  \centering
  \resizebox{\columnwidth}{!}{ 
  \begin{tabular}{c|cc|c|c|c}
    \toprule
    \multirow{2}{*}[-0.5ex]{Model} & \multicolumn{2}{c|}{Image Prompt} & \multirow{2}{*}[-0.5ex]{Text Prompt} & \multicolumn{2}{c}{GP$\uparrow$}\\
    \cmidrule(lr{0.1cm}){2-3}
    \cmidrule(lr{0.1cm}){5-5}
    \cmidrule(lr{0.1cm}){6-6}
     & Train & Eval &  & Val Seen  & Val Unseen   \\
    \midrule
    \multirow{5}{*}{HAMT}   
                            & $\times$ & $\times$  & $\checkmark$        & 6.30 & 2.43   \\
                            & $\times$ & $\times$  & $\checkmark^{*}$     & 9.80 & 2.52   \\
                            & $\checkmark$  & $\times$ & $\checkmark$    & 7.03 & 2.49   \\
                            & $\checkmark$ & $\checkmark$ & $\times$  & 8.83 & 2.29   \\
                            & $\checkmark$ & $\checkmark$ & $\checkmark$ & \textbf{10.92} & \textbf{3.11}   \\
    \midrule
    \multirow{5}{*}{DUET}   
                            & $\times$ & $\times$  & $\checkmark$        & 6.19  & 3.00  \\
                            & $\times$ & $\times$  & $\checkmark^{*}$      & 9.61  & 3.12   \\
                            & $\checkmark$  & $\times$  & $\checkmark$   & 6.27  & 3.57  \\
                            & $\checkmark$ & $\checkmark$ & $\times$  & 7.89  & 3.12   \\
                            & $\checkmark$ & $\checkmark$ & $\checkmark$ & \textbf{10.09} & \textbf{4.13}   \\
    \bottomrule
  \end{tabular}
  }
\caption{Navigation performance on the CVDN-MP. When visual prompts are used, MPF is integrated. * means with captions.}
\label{tab:cvdn}
\end{table}

\subsection{Ablation Study}

\paragraph{Proportion of Augmented Data.}
In \cref{tab:aug}, we evaluate the impact of the ratio $\gamma$ of ControlNet-generated augmented data to original data during training. 
As $\gamma$ increases, performance in the val seen split gradually declines. 
While in the val unseen split, it initially improves, peaking at $\gamma = 0.2$, before decreasing. 
This trend is intuitive, as augmented data mitigates over-fitting to familiar landmarks and improves adaptability to new environments. 
However, excessive augmented data can create a disparity between prompts and observations and confuse agents. 

\begin{table}[t]
\centering
\begin{tabular}{c|c|cc|cc}
\toprule
\multirow{2}{*}[-0.5ex]{\#} &  \multirow{2}{*}[-0.5ex]{$\gamma$} &         \multicolumn{2}{c|}{\textbf{Val Seen}} & \multicolumn{2}{c}{\textbf{Val Unseen}}\\
\cmidrule(lr{0.2cm}){3-4}
\cmidrule(lr{0.2cm}){5-6}
 &               & SR$\uparrow$   & SPL$\uparrow$   & SR$\uparrow$   & SPL$\uparrow$   \\
\hline
1 & 0.0               & \textbf{69.2}   & \textbf{65.5}         & 60.0   & 56.3     \\
2 & 0.1               & 68.4   & 63.9         & 60.2   & 55.2       \\
3 & 0.2               & 67.6   & 63.5      & \textbf{61.8}   & \textbf{57.2}     \\
4 & 0.5               & 66.8   & 62.6      & 59.8   & 55.5     \\
5 & 0.8               & 66.4   & 62.2       & 59.6   & 55.1      \\
6 & 1.0               & 64.0   & 59.9     & 58.8   & 54.6      \\
\bottomrule
\end{tabular}
\caption{Results with different proportions of augmented data.}
\label{tab:aug}
\end{table}

\paragraph{Position Encoding.}
\cref{tab:pe} presents the navigation performance of the MPF module with different position encodings in the Aligned setting of RxR-MP.
To distinguish them, we refer to the first as visual position encoding (VPE), as it is only applied to the visual prompts, and the second as multi-modal position encoding (MPE), as it is applied to the multi-modal tokens.
Both types of position encoding are beneficial to agents when dealing with multi-modal instructions. 
Since MPE implies VPE, experiment \#3 in \cref{tab:pe} outperforms \#2.
Combining these two position encodings in experiment \#4 results in the best performance, supporting our assertion of their distinct functions.
\begin{table}[t]
\centering
\begin{tabular}{c|cc|cc|cc}
\toprule
\multirow{2}{*}[-0.5ex]{\#} & \multirow{2}{*}[-0.5ex]{VPE} & \multirow{2}{*}[-0.5ex]{MPE} & \multicolumn{2}{c|}{\textbf{Val Seen}} & \multicolumn{2}{c}{\textbf{Val Unseen}} \\
\cmidrule(lr{0.2cm}){4-5}
\cmidrule(lr{0.2cm}){6-7}
 &         &           & SR$\uparrow$   & SPL$\uparrow$  & SR$\uparrow$   & SPL$\uparrow$   \\
\hline
1 & $\times$         & $\times$             & 67.6   & 63.3       & 58.7   & 53.8   \\
2 & $\checkmark$   & $\times$               & 68.0   & 63.4       & 59.0   & 54.4   \\
3 & $\times$      & $\checkmark$             & 67.4   & 62.9       & 59.3   & 55.2   \\
4 & $\checkmark$    & $\checkmark$             & \textbf{69.2}   & \textbf{65.5}       & \textbf{60.0}   & \textbf{56.3}   \\
\bottomrule
\end{tabular}
\caption{Navigation performance of MPF with different position encodings on the Aligned setting of RxR-MP.}
\label{tab:pe}
\end{table}

\subsection{Pre-explore Setting of VLN}
Most VLN works focus on the single-run setting, where agents navigate in an environment once without prior knowledge or multiple trials.
However, the pre-explore setting, where agents familiarize the environment before navigation, is also crucial, especially for household robots that operate persistently in an environment~\cite{krantz2023iterative}.
VLN-MP allows existing VLN models to be applied to this pre-explore setting with only one more trial and enhances performance without additional training. 
Specifically, the agent first follows the textual instruction to generate a pseudo path and then employs our pipeline to extract landmark images from the visited nodes and their neighbors to form a multi-modal instruction for navigation.
\cref{tab:pre} show that this method can significantly improve navigation performance in RxR, with a notable 6.9\% and 2.5\% SR increase in seen and unseen scenarios compared to the original HAMT, emphasizing the applicability of VLN-MP in traditional VLN tasks.

\begin{table}[t]
\centering
\begin{tabular}{p{30pt}|c|cc|cc}
\toprule
 \multirow{2}{*}[-0.5ex]{Model} & \multirow{2}{*}[-0.5ex]{Pre-explore} & \multicolumn{2}{c|}{\textbf{Val Seen}} & \multicolumn{2}{c}{\textbf{Val Unseen}}\\
\cmidrule(lr{0.1cm}){3-4}
\cmidrule(lr{0.1cm}){5-6}
  &  & SR$\uparrow$   & SPL$\uparrow$     & SR$\uparrow$   & SPL$\uparrow$    \\ \hline
HAMT      & $\times$            & 59.4  & 58.9      & 56.5   &  \textbf{56.0}        \\
Ours     & $\times$                  & 64.6  & 60.8       & 57.6   & 53.5         \\
Ours      & $\checkmark$                  & \textbf{66.3}  & \textbf{62.1}       & \textbf{59.0}   &  55.4           \\
\bottomrule
\end{tabular}
\caption{Comparison of HAMT and our HAMT+MPF agents on RxR dataset under different settings.}
\label{tab:pre}
\end{table}

\section{Conclusion}
In this paper, we propose the novel VLN-MP task, which enhances agent navigation by integrating visual prompts into textual instructions. 
VLN-MP extends traditional VLN by ensuring full backward compatibility and demonstrating adaptability to various visual prompts.
We establish the first benchmark for VLN-MP, including a pipeline for converting textual instructions into multi-modal forms, four datasets for different downstream tasks, and a novel MPF module to process multi-modal instructions efficiently.
We conduct extensive experiments to evaluate our proposed components and provide empirical insights regarding the usage of visual prompts. 
Notably, VLN-MP allows agents to operate in a pre-explore setting, achieving enhanced results compared to traditional VLN. 
We believe that VLN-MP and our benchmark will broaden the scope of VLN in real-world applications and open up new research opportunities in the field. 

\newpage
\section*{Acknowledgements}
This work was partially supported by ARC DECRA grant (DE200101610) and CSIRO's Science Leader project R-91559.

\bibliographystyle{named}
\bibliography{ijcai24}

\begin{thebibliography}{}

\bibitem[\protect\citeauthoryear{Alayrac \bgroup \em et al.\egroup }{2022}]{alayrac2022flamingo}
Jean-Baptiste Alayrac, Jeff Donahue, Pauline Luc, Antoine Miech, Iain Barr, Yana Hasson, Karel Lenc, Arthur Mensch, Katherine Millican, Malcolm Reynolds, et~al.
\newblock Flamingo: a visual language model for few-shot learning.
\newblock {\em NeurIPS}, 35:23716--23736, 2022.

\bibitem[\protect\citeauthoryear{Anderson \bgroup \em et al.\egroup }{2018a}]{anderson2018evaluation}
Peter Anderson, Angel Chang, Devendra~Singh Chaplot, Alexey Dosovitskiy, Saurabh Gupta, Vladlen Koltun, Jana Kosecka, Jitendra Malik, Roozbeh Mottaghi, Manolis Savva, et~al.
\newblock On evaluation of embodied navigation agents.
\newblock {\em arXiv preprint arXiv:1807.06757}, 2018.

\bibitem[\protect\citeauthoryear{Anderson \bgroup \em et al.\egroup }{2018b}]{anderson2018vision}
Peter Anderson, Qi~Wu, Damien Teney, Jake Bruce, Mark Johnson, Niko S{\"u}nderhauf, Ian Reid, Stephen Gould, and Anton Van Den~Hengel.
\newblock Vision-and-language navigation: Interpreting visually-grounded navigation instructions in real environments.
\newblock In {\em CVPR}, pages 3674--3683, 2018.

\bibitem[\protect\citeauthoryear{Berg \bgroup \em et al.\egroup }{2020}]{berg2020grounding}
Matthew Berg, Deniz Bayazit, Rebecca Mathew, Ariel Rotter-Aboyoun, Ellie Pavlick, and Stefanie Tellex.
\newblock Grounding language to landmarks in arbitrary outdoor environments.
\newblock In {\em ICRA}, pages 208--215, 2020.

\bibitem[\protect\citeauthoryear{Brown \bgroup \em et al.\egroup }{2020}]{brown2020language}
Tom Brown, Benjamin Mann, Nick Ryder, Melanie Subbiah, Jared~D Kaplan, Prafulla Dhariwal, Arvind Neelakantan, Pranav Shyam, Girish Sastry, Amanda Askell, et~al.
\newblock Language models are few-shot learners.
\newblock {\em NeurIPS}, 33:1877--1901, 2020.

\bibitem[\protect\citeauthoryear{Chang \bgroup \em et al.\egroup }{2017}]{chang2017matterport3d}
Angel Chang, Angela Dai, Thomas Funkhouser, Maciej Halber, Matthias Niebner, Manolis Savva, Shuran Song, Andy Zeng, and Yinda Zhang.
\newblock Matterport3d: Learning from rgb-d data in indoor environments.
\newblock In {\em 3DV}, pages 667--676, 2017.

\bibitem[\protect\citeauthoryear{Chen \bgroup \em et al.\egroup }{2021}]{chen2021history}
Shizhe Chen, Pierre-Louis Guhur, Cordelia Schmid, and Ivan Laptev.
\newblock History aware multimodal transformer for vision-and-language navigation.
\newblock {\em NeurIPS}, 34:5834--5847, 2021.

\bibitem[\protect\citeauthoryear{Chen \bgroup \em et al.\egroup }{2022}]{chen2022think}
Shizhe Chen, Pierre-Louis Guhur, Makarand Tapaswi, Cordelia Schmid, and Ivan Laptev.
\newblock Think global, act local: Dual-scale graph transformer for vision-and-language navigation.
\newblock In {\em CVPR}, pages 16537--16547, 2022.

\bibitem[\protect\citeauthoryear{Cui \bgroup \em et al.\egroup }{2023}]{cui2023grounded}
Yibo Cui, Liang Xie, Yakun Zhang, Meishan Zhang, Ye~Yan, and Erwei Yin.
\newblock Grounded entity-landmark adaptive pre-training for vision-and-language navigation.
\newblock In {\em ICCV}, pages 12043--12053, 2023.

\bibitem[\protect\citeauthoryear{Devlin \bgroup \em et al.\egroup }{2019}]{kenton2019bert}
Jacob Devlin, Ming-Wei Chang, Kenton Lee, and Kristina Toutanova.
\newblock Bert: Pre-training of deep bidirectional transformers for language understanding.
\newblock In {\em Proceedings of NAACL-HLT}, pages 4171--4186, 2019.

\bibitem[\protect\citeauthoryear{Gu \bgroup \em et al.\egroup }{2022}]{gu2022towards}
Geonmo Gu, Byungsoo Ko, SeoungHyun Go, Sung-Hyun Lee, Jingeun Lee, and Minchul Shin.
\newblock Towards light-weight and real-time line segment detection.
\newblock In {\em AAAI}, volume~36, pages 726--734, 2022.

\bibitem[\protect\citeauthoryear{He \bgroup \em et al.\egroup }{2021}]{he2021landmark}
Keji He, Yan Huang, Qi~Wu, Jianhua Yang, Dong An, Shuanglin Sima, and Liang Wang.
\newblock Landmark-rxr: Solving vision-and-language navigation with fine-grained alignment supervision.
\newblock {\em NeurIPS}, 34:652--663, 2021.

\bibitem[\protect\citeauthoryear{Hong \bgroup \em et al.\egroup }{2020}]{hong2020language}
Yicong Hong, Cristian Rodriguez, Yuankai Qi, Qi~Wu, and Stephen Gould.
\newblock Language and visual entity relationship graph for agent navigation.
\newblock {\em NeurIPS}, 33:7685--7696, 2020.

\bibitem[\protect\citeauthoryear{Huang \bgroup \em et al.\egroup }{2022}]{huang2022language}
Wenlong Huang, Pieter Abbeel, Deepak Pathak, and Igor Mordatch.
\newblock Language models as zero-shot planners: Extracting actionable knowledge for embodied agents.
\newblock In {\em ICML}, pages 9118--9147, 2022.

\bibitem[\protect\citeauthoryear{Ilharco \bgroup \em et al.\egroup }{2019}]{ilharco2019general}
Gabriel Ilharco, Vihan Jain, Alexander Ku, Eugene Ie, and Jason Baldridge.
\newblock General evaluation for instruction conditioned navigation using dynamic time warping.
\newblock {\em arXiv preprint arXiv:1907.05446}, 2019.

\bibitem[\protect\citeauthoryear{Jain \bgroup \em et al.\egroup }{2019}]{jain2019stay}
Vihan Jain, Gabriel Magalhaes, Alexander Ku, Ashish Vaswani, Eugene Ie, and Jason Baldridge.
\newblock Stay on the path: Instruction fidelity in vision-and-language navigation.
\newblock In {\em ACL}, pages 1862--1872, 2019.

\bibitem[\protect\citeauthoryear{Jiang \bgroup \em et al.\egroup }{2022}]{jiang2022vima}
Yunfan Jiang, Agrim Gupta, Zichen Zhang, Guanzhi Wang, Yongqiang Dou, Yanjun Chen, Li~Fei-Fei, Anima Anandkumar, Yuke Zhu, and Linxi Fan.
\newblock {VIMA}: General robot manipulation with multimodal prompts.
\newblock In {\em NeurIPS 2022 Foundation Models for Decision Making Workshop}, 2022.

\bibitem[\protect\citeauthoryear{Khattak \bgroup \em et al.\egroup }{2023}]{khattak2023maple}
Muhammad~Uzair Khattak, Hanoona Rasheed, Muhammad Maaz, Salman Khan, and Fahad~Shahbaz Khan.
\newblock Maple: Multi-modal prompt learning.
\newblock In {\em CVPR}, pages 19113--19122, 2023.

\bibitem[\protect\citeauthoryear{Krantz \bgroup \em et al.\egroup }{2020}]{krantz2020beyond}
Jacob Krantz, Erik Wijmans, Arjun Majumdar, Dhruv Batra, and Stefan Lee.
\newblock Beyond the nav-graph: Vision-and-language navigation in continuous environments.
\newblock In {\em ECCV}, pages 104--120, 2020.

\bibitem[\protect\citeauthoryear{Krantz \bgroup \em et al.\egroup }{2023}]{krantz2023iterative}
Jacob Krantz, Shurjo Banerjee, Wang Zhu, Jason Corso, Peter Anderson, Stefan Lee, and Jesse Thomason.
\newblock Iterative vision-and-language navigation.
\newblock In {\em CVPR}, pages 14921--14930, 2023.

\bibitem[\protect\citeauthoryear{Ku \bgroup \em et al.\egroup }{2020}]{ku2020room}
Alexander Ku, Peter Anderson, Roma Patel, Eugene Ie, and Jason Baldridge.
\newblock Room-across-room: Multilingual vision-and-language navigation with dense spatiotemporal grounding.
\newblock In {\em EMNLP}, pages 4392--4412, 2020.

\bibitem[\protect\citeauthoryear{Li \bgroup \em et al.\egroup }{2022a}]{li2022clear}
Jialu Li, Hao Tan, and Mohit Bansal.
\newblock Clear: Improving vision-language navigation with cross-lingual, environment-agnostic representations.
\newblock In {\em NAACL}, pages 633--649, 2022.

\bibitem[\protect\citeauthoryear{Li \bgroup \em et al.\egroup }{2022b}]{li2022envedit}
Jialu Li, Hao Tan, and Mohit Bansal.
\newblock Envedit: Environment editing for vision-and-language navigation.
\newblock In {\em CVPR}, pages 15407--15417, 2022.

\bibitem[\protect\citeauthoryear{Li \bgroup \em et al.\egroup }{2022c}]{li2022grounded}
Liunian~Harold Li, Pengchuan Zhang, Haotian Zhang, Jianwei Yang, Chunyuan Li, Yiwu Zhong, Lijuan Wang, Lu~Yuan, Lei Zhang, Jenq-Neng Hwang, et~al.
\newblock Grounded language-image pre-training.
\newblock In {\em CVPR}, pages 10965--10975, 2022.

\bibitem[\protect\citeauthoryear{Li \bgroup \em et al.\egroup }{2023}]{li2023blip}
Junnan Li, Dongxu Li, Silvio Savarese, and Steven Hoi.
\newblock Blip-2: Bootstrapping language-image pre-training with frozen image encoders and large language models.
\newblock In {\em ICML}, pages 19730--19742, 2023.

\bibitem[\protect\citeauthoryear{Lin}{2004}]{lin2004rouge}
Chin-Yew Lin.
\newblock Rouge: A package for automatic evaluation of summaries.
\newblock In {\em Text summarization branches out}, pages 74--81, 2004.

\bibitem[\protect\citeauthoryear{Liu \bgroup \em et al.\egroup }{2021}]{liu2021vision}
Chong Liu, Fengda Zhu, Xiaojun Chang, Xiaodan Liang, Zongyuan Ge, and Yi-Dong Shen.
\newblock Vision-language navigation with random environmental mixup.
\newblock In {\em ICCV}, pages 1644--1654, 2021.

\bibitem[\protect\citeauthoryear{Liu \bgroup \em et al.\egroup }{2023a}]{liu2023grounding}
Shilong Liu, Zhaoyang Zeng, Tianhe Ren, Feng Li, Hao Zhang, Jie Yang, Chunyuan Li, Jianwei Yang, Hang Su, Jun Zhu, et~al.
\newblock Grounding dino: Marrying dino with grounded pre-training for open-set object detection.
\newblock {\em arXiv preprint arXiv:2303.05499}, 2023.

\bibitem[\protect\citeauthoryear{Liu \bgroup \em et al.\egroup }{2023b}]{liu2023aerialvln}
Shubo Liu, Hongsheng Zhang, Yuankai Qi, Peng Wang, Yanning Zhang, and Qi~Wu.
\newblock Aerialvln: Vision-and-language navigation for uavs.
\newblock In {\em ICCV}, pages 15384--15394, 2023.

\bibitem[\protect\citeauthoryear{Majumdar \bgroup \em et al.\egroup }{2020}]{majumdar2020improving}
Arjun Majumdar, Ayush Shrivastava, Stefan Lee, Peter Anderson, Devi Parikh, and Dhruv Batra.
\newblock Improving vision-and-language navigation with image-text pairs from the web.
\newblock In {\em ECCV}, pages 259--274, 2020.

\bibitem[\protect\citeauthoryear{OpenAI}{2023}]{openai2023gpt4}
OpenAI.
\newblock Gpt-4 technical report.
\newblock {\em arXiv preprint arXiv:2303.08774}, 2023.

\bibitem[\protect\citeauthoryear{Qi \bgroup \em et al.\egroup }{2020}]{qi2020reverie}
Yuankai Qi, Qi~Wu, Peter Anderson, Xin Wang, William~Yang Wang, Chunhua Shen, and Anton van~den Hengel.
\newblock Reverie: Remote embodied visual referring expression in real indoor environments.
\newblock In {\em CVPR}, pages 9982--9991, 2020.

\bibitem[\protect\citeauthoryear{Qiao \bgroup \em et al.\egroup }{2023a}]{qiao2023hop+}
Yanyuan Qiao, Yuankai Qi, Yicong Hong, Zheng Yu, Peng Wang, and Qi~Wu.
\newblock Hop+: History-enhanced and order-aware pre-training for vision-and-language navigation.
\newblock {\em TPAMI}, 2023.

\bibitem[\protect\citeauthoryear{Qiao \bgroup \em et al.\egroup }{2023b}]{qiao2023vln}
Yanyuan Qiao, Zheng Yu, and Qi~Wu.
\newblock Vln-petl: Parameter-efficient transfer learning for vision-and-language navigation.
\newblock In {\em ICCV}, pages 15443--15452, 2023.

\bibitem[\protect\citeauthoryear{Radford \bgroup \em et al.\egroup }{2021}]{radford2021learning}
Alec Radford, Jong~Wook Kim, Chris Hallacy, Aditya Ramesh, Gabriel Goh, Sandhini Agarwal, Girish Sastry, Amanda Askell, Pamela Mishkin, Jack Clark, et~al.
\newblock Learning transferable visual models from natural language supervision.
\newblock In {\em ICML}, pages 8748--8763, 2021.

\bibitem[\protect\citeauthoryear{Shen \bgroup \em et al.\egroup }{2022}]{shen2021much}
Sheng Shen, Liunian~Harold Li, Hao Tan, Mohit Bansal, Anna Rohrbach, Kai-Wei Chang, Zhewei Yao, and Kurt Keutzer.
\newblock How much can clip benefit vision-and-language tasks?
\newblock {\em ICLR}, 2022.

\bibitem[\protect\citeauthoryear{Thomason \bgroup \em et al.\egroup }{2020}]{thomason2020vision}
Jesse Thomason, Michael Murray, Maya Cakmak, and Luke Zettlemoyer.
\newblock Vision-and-dialog navigation.
\newblock In {\em CoRL}, pages 394--406, 2020.

\bibitem[\protect\citeauthoryear{Touvron \bgroup \em et al.\egroup }{2023}]{touvron2023llama}
Hugo Touvron, Thibaut Lavril, Gautier Izacard, Xavier Martinet, Marie-Anne Lachaux, Timoth{\'e}e Lacroix, Baptiste Rozi{\`e}re, Naman Goyal, Eric Hambro, Faisal Azhar, et~al.
\newblock Llama: Open and efficient foundation language models.
\newblock {\em arXiv preprint arXiv:2302.13971}, 2023.

\bibitem[\protect\citeauthoryear{Vaswani \bgroup \em et al.\egroup }{2017}]{vaswani2017attention}
Ashish Vaswani, Noam Shazeer, Niki Parmar, Jakob Uszkoreit, Llion Jones, Aidan~N Gomez, {\L}ukasz Kaiser, and Illia Polosukhin.
\newblock Attention is all you need.
\newblock {\em NeurIPS}, 30, 2017.

\bibitem[\protect\citeauthoryear{Wang \bgroup \em et al.\egroup }{2022}]{wang2022less}
Su~Wang, Ceslee Montgomery, Jordi Orbay, Vighnesh Birodkar, Aleksandra Faust, Izzeddin Gur, Natasha Jaques, Austin Waters, Jason Baldridge, and Peter Anderson.
\newblock Less is more: Generating grounded navigation instructions from landmarks.
\newblock In {\em CVPR}, pages 15428--15438, 2022.

\bibitem[\protect\citeauthoryear{Wang \bgroup \em et al.\egroup }{2023a}]{wang2023dual}
Liuyi Wang, Zongtao He, Jiagui Tang, Ronghao Dang, Naijia Wang, Chengju Liu, and Qijun Chen.
\newblock A dual semantic-aware recurrent global-adaptive network for vision-and-language navigation.
\newblock In {\em IJCAI}, pages 1479--1487, 2023.

\bibitem[\protect\citeauthoryear{Wang \bgroup \em et al.\egroup }{2023b}]{wang2023scaling}
Zun Wang, Jialu Li, Yicong Hong, Yi~Wang, Qi~Wu, Mohit Bansal, Stephen Gould, Hao Tan, and Yu~Qiao.
\newblock Scaling data generation in vision-and-language navigation.
\newblock In {\em ICCV}, pages 12009--12020, 2023.

\bibitem[\protect\citeauthoryear{Yesiltepe \bgroup \em et al.\egroup }{2021}]{yesiltepe2021landmarks}
Demet Yesiltepe, Ruth Conroy~Dalton, and Ayse Ozbil~Torun.
\newblock Landmarks in wayfinding: a review of the existing literature.
\newblock {\em Cognitive processing}, 22:369--410, 2021.

\bibitem[\protect\citeauthoryear{Zhang \bgroup \em et al.\egroup }{2021}]{zhang2021diagnosing}
Yubo Zhang, Hao Tan, and Mohit Bansal.
\newblock Diagnosing the environment bias in vision-and-language navigation.
\newblock In {\em IJCAI}, pages 890--897, 2021.

\bibitem[\protect\citeauthoryear{Zhang \bgroup \em et al.\egroup }{2023}]{zhang2023adding}
Lvmin Zhang, Anyi Rao, and Maneesh Agrawala.
\newblock Adding conditional control to text-to-image diffusion models.
\newblock In {\em ICCV}, pages 3836--3847, 2023.

\bibitem[\protect\citeauthoryear{Zhou \bgroup \em et al.\egroup }{2024}]{zhou2024navgpt}
Gengze Zhou, Yicong Hong, and Qi~Wu.
\newblock Navgpt: Explicit reasoning in vision-and-language navigation with large language models.
\newblock In {\em AAAI}, pages 7641--7649, 2024.

\bibitem[\protect\citeauthoryear{Zhu \bgroup \em et al.\egroup }{2020}]{zhu2020vision}
Fengda Zhu, Yi~Zhu, Xiaojun Chang, and Xiaodan Liang.
\newblock Vision-language navigation with self-supervised auxiliary reasoning tasks.
\newblock In {\em CVPR}, pages 10012--10022, 2020.

\bibitem[\protect\citeauthoryear{Zhu \bgroup \em et al.\egroup }{2021}]{zhu2021soon}
Fengda Zhu, Xiwen Liang, Yi~Zhu, Qizhi Yu, Xiaojun Chang, and Xiaodan Liang.
\newblock Soon: Scenario oriented object navigation with graph-based exploration.
\newblock In {\em CVPR}, pages 12689--12699, 2021.

\bibitem[\protect\citeauthoryear{Zhu \bgroup \em et al.\egroup }{2023}]{zhu2023vision}
Fengda Zhu, Vincent~CS Lee, Xiaojun Chang, and Xiaodan Liang.
\newblock Vision language navigation with knowledge-driven environmental dreamer.
\newblock In {\em IJCAI}, pages 1840--1848, 2023.

\end{thebibliography}

\end{document}